\documentclass{article}
\usepackage{amssymb}
\usepackage{amsmath}
\usepackage{graphicx}
\usepackage{algorithm}
\usepackage{algpseudocode}
\usepackage{wrapfig}
\usepackage{bm}
\usepackage{multirow}
\usepackage[table]{xcolor}
\usepackage{tikzducks}

\usepackage{multirow}
\usepackage{adjustbox}
\usepackage{caption}
\usepackage[preprint]{corl_2026} 
\usepackage{booktabs}

\usepackage{subcaption}

\usepackage{makecell}
\usepackage{pifont}

\title{
SceneBot: Contact-Prompted General Humanoid Whole Body Tracking with Scene-Interaction
}

\author{
  Sirui Chen\\
  Stanford, Amazon FAR\\
  United States \\
  \texttt{ericcsr@stanford.edu} \\
  \And
  Shibo Zhao \\
  Amazon FAR\\
  United States \\
  \texttt{shiboxx@amazon.com} \\
  \And
  Zhen Wu \\
  Amazon FAR\\
  United States \\
  \texttt{zhenwuuu@amazon.com} \\
  \AND
  Jiaman Li \\
  Amazon FAR\\
  United States \\
  \texttt{jiamanli@amazon.com} \\
  \And
  Guanya Shi$^\dagger$ \\
  CMU, Amazon FAR\\
  United States \\
  \texttt{guanyas@amazon.com} \\
  \And
  C. Karen Liu$^\dagger$ \\
  Stanford, Amazon FAR\\
  United States \\
  \texttt{karenliu@cs.stanford.edu} \\
}

\newcommand{\cmark}{\checkmark}
\newcommand{\xmark}{\ding{55}}

\begin{document}
\maketitle

\begin{figure}[h]
  \begin{center}
  \includegraphics[width=1.0\linewidth]{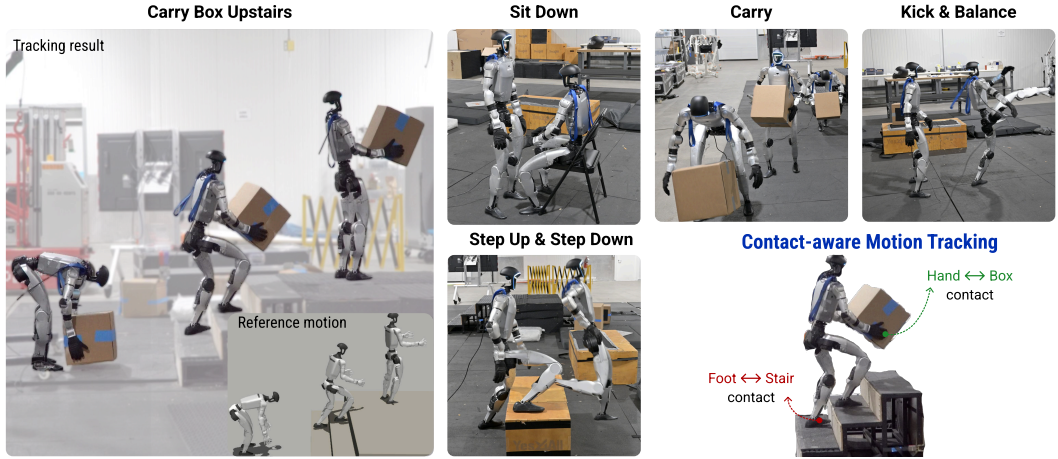}
  \end{center}
  \vspace{-5pt}
  \caption{SceneBot enables a single motion tracking policy to accurately achieve free-form locomotion, uneven terrain traversal, and object manipulation.}
  \vspace{-15pt}
  \label{fig:teaser}
\end{figure}
\begin{abstract}

Current humanoid reinforcement-learning policies excel at free-space motions but struggle with contact-rich tasks, as pure kinematic tracking cannot resolve the physical ambiguities of interacting with objects and uneven terrain. To address this, we introduce SceneBot, a unified motion-tracking framework capable of handling freespace locomotion, terrain traversal, and whole-body manipulation. SceneBot sconditions a single policy on both reference motions and per-link contact labels, explicitly defining expected environmental interactions. To overcome the lack of annotated interaction data, we propose a hindsight scene reconstruction approach that infers scene-interaction graphs from retargeted human motion. Trained on 7.5 hours of this reconstructed, contact-rich data, SceneBot successfully generalizes to unseen motions and environments. Our results demonstrate that SceneBot is the first general framework to seamlessly unify free-space and contact-rich behaviors—executing complex, long-horizon tasks like carrying a box upstairs and establishing contact conditioning as a powerful interface for humanoid control. All code and data will be open-sourced. More demos and informations are available at: \url{https://ericcsr.github.io/scenebot/}

\end{abstract}

\section{Introduction}
\label{sec:intro}

Physical intelligence requires robots to intentionally leverage contacts with the world for support, manipulation and locomotion.  Consider a household robot picking up a basket, climbing stairs while maintaining balance, and placing the basket near a washing machine. Accomplishing such a contact-rich task requires coordinated interaction with both objects and terrain over long horizons.

Recent progress in humanoid control has been driven by reinforcement-learning policies trained to track large collections of reference motions. These general motion trackers can execute diverse free-space behaviors such as walking, running, dancing, and kicking with a single policy. However, their success has largely been limited to environments where the reference motion can be followed without reasoning about external contact. Once the robot interacts with large objects or non-flat terrain, kinematics alone becomes ambiguous. Moving both wrists to the sides of a box may correspond to reaching toward the box, lightly touching it, or applying sufficient force to lift it. Similarly, placing a foot near the edge of a stair does not guarantee that the robot generates sufficient ground reaction force to step upward. Thus, successful task execution depends not only on reaching desired poses, but also on knowing which body parts should establish, maintain, and utilize contact forces with the environment.

We introduce SceneBot, a contact-conditioning motion tracking framework that enables a single policy to track humanoid free-form locomotion, terrain traversal (e.g. stairs), and whole-body manipulation. SceneBot conditions a general, whole-body policy on both reference motion and \emph{per-link contact labels}. The reference motion specifies desired kinematic behavior, while contact labels explicitly indicate which body links should expect and exploit contact forces from different parts of the scene. Compared to existing motion trackers, SceneBot asks the high-level decision maker to provide one additional command: which robot links should actively establish contact with the scene. However, these labels are defined only on the robot body, not on scene objects or terrain. This design keeps SceneBot as a low-level motion tracking policy, with a minimal extension for scene interaction without turning into a vision-based task policy.

Training such a contact-aware controller presents a major data challenge: it requires motion trajectories paired with corresponding object geometry, terrain geometry, and contact annotations. Unfortunately, large-scale datasets containing both motion and scene interaction information are scarce, and scalable scene-aware retargeting remains difficult. SceneBot addresses this data bottleneck through hindsight scene reconstruction. Given a retargeted robot motion, we infer a scene-interaction graph that identifies which robot links should contact terrain or movable objects over time. We then reconstruct plausible terrain and object assets consistent with those inferred contacts. This turns ordinary human motion data into contact-rich robot-scene interaction data suitable for training.

We show that Scenebot is, to the best of our knowledge, the first general motion tracking framework capable of handling free-space locomotion, uneven terrain traversal, and whole-body object manipulation within a single policy. Scenebot generalizes to unseen motions and unseen environments, while maintaining performance comparable to state-of-the-art motion trackers in free-space settings. We further demonstrate long-horizon tasks requiring simultaneous terrain and object interaction, such as picking up a large box, carrying it, and walking upstairs. Our results show that contact conditioning provides an effective interface for extending general humanoid motion tracking from free-space motion to contact-rich scene interaction.

\section{Related Works}
\label{sec:related}
\subsection{General motion tracking}
Recent RL-based motion tracking has evolved from task-specific policies \cite{beyondmimic, deepmimic, asap} to unified controllers capable of tracking diverse trajectories \cite{sonic, gmt, any2track, twist, gentlehumanoid, chip}. Supported by large-scale datasets \cite{sonic, amass}, scalable retargeting \cite{soma, gmr}, and GPU simulators \cite{isaaclab, mjlab}, these generalized policies serve as robust low-level controllers for downstream tasks like teleoperation and Vision-Language-Action architectures \cite{twist, twist2, sonic, chip}. However, a critical gap remains: current pipelines primarily target free-space movement. Existing datasets and retargeting methods fundamentally lack the scene-awareness required for complex human-environment interactions.

\subsection{Terrain aware whole body control}
Mastering complex terrain traversal has shifted from model-based footstep planning and MPC \cite{deits2014footstep, atlas} to RL policies utilizing root velocity commands and gait-based reward shaping \cite{gallant, rpl, ame2}. Agile maneuvers, such as parkour, have been achieved through scene-aware retargeting \cite{omniretarget}, policy distillation \cite{php}, and AMP-style rewards \cite{amp, intrinsic}. Despite these advances, most terrain traversal policies rely exclusively on root velocity control, severely limiting their utility as general-purpose whole-body controllers for loco-manipulation. While recent works \cite{parc, zhang2026learning} attempt to bridge this gap by co-training controllers with terrain-aware kinematic generators, their reliance on kinematically generated data significantly constrains dexterous upper-body movement.

\subsection{Humanoid object interaction}
Beyond locomotion, humanoid object manipulation has largely been confined to single-handed, lightweight pick-and-place tasks \cite{groot, sonic, psi0}. For bimanual grasping of heavy objects, existing impedance controllers \cite{chip} still depend on high-level policies or human teleoperators to manually regulate force. Although recent methods leverage human-object datasets \cite{omomo, humoto} to achieve single-trajectory tracking for large objects \cite{hdmi, omniretarget, hihi}, the field lacks a general-purpose, whole-body controller capable of zero-shot deployment for complex manipulation. To address this, our work presents a generalized motion tracking controller designed for zero-shot deployment, seamlessly enabling both unseen terrain traversal and robust object interaction.

\begin{table}[htbp]
\centering
\resizebox{0.8\textwidth}{!}{
\begin{tabular}{l c c c c}
\toprule
& \makecell{\textbf{Tracking} \\ \textbf{Single Motion}} & \makecell{\textbf{Tracking} \\ \textbf{Diverse Motion}} & \makecell{\textbf{Terrain} \\ \textbf{Interaction}} & \makecell{\textbf{Object} \\ \textbf{Interaction}} \\
\midrule
BeyondMimic \textcolor{blue}{\cite{beyondmimic}}  & \cmark & \xmark & \xmark & \xmark \\
OmniRetarget \textcolor{blue}{\cite{omniretarget}} & \cmark & \xmark & \cmark & \cmark \\
HDMI \textcolor{blue}{\cite{hdmi}}         & \cmark & \xmark & \xmark & \cmark \\
ResMimic \textcolor{blue}{\cite{resmimic}}     & \cmark & \xmark & \xmark & \cmark \\
SONIC \textcolor{blue}{\cite{sonic}}        & \cmark & \cmark & \xmark & \xmark \\
CHIP \textcolor{blue}{\cite{chip}}         & \cmark & \cmark & \xmark & \cmark \\
TWIST \textcolor{blue}{\cite{twist}}        & \cmark & \cmark & \xmark & \xmark \\
Any2Track \textcolor{blue}{\cite{any2track}}    & \cmark & \cmark & \xmark & \xmark \\
PHP \textcolor{blue}{\cite{php}}          & \xmark & \xmark & \cmark & \xmark \\
RPL \textcolor{blue}{\cite{rpl}}          & \xmark & \xmark & \cmark & \xmark \\
WoCoCo \textcolor{blue}{\cite{wococo}}       & \xmark & \xmark & \cmark & \cmark \\
Gallent \textcolor{blue}{\cite{gallant}}      & \xmark & \xmark & \cmark & \xmark \\
HikeWild \textcolor{blue}{\cite{intrinsic}}     & \xmark & \xmark & \cmark & \xmark \\
\midrule
\textbf{Scenebot}                   & \cmark & \cmark & \cmark & \cmark \\
\bottomrule
\end{tabular}
}
\vspace{5pt}
\caption{\textbf{Comparison of whole body controllers.} \textbf{Scenebot} can track diverse reference motion and contact labels to achieve different scene interactions such as terrain interaction and carrying large objects.}
\label{tab:robot_data_generation}
\vspace{-1.5pt}
\end{table}



\section{Method}
\label{sec:method}
\begin{figure}[h]
  \centering
  \includegraphics[width=\linewidth]{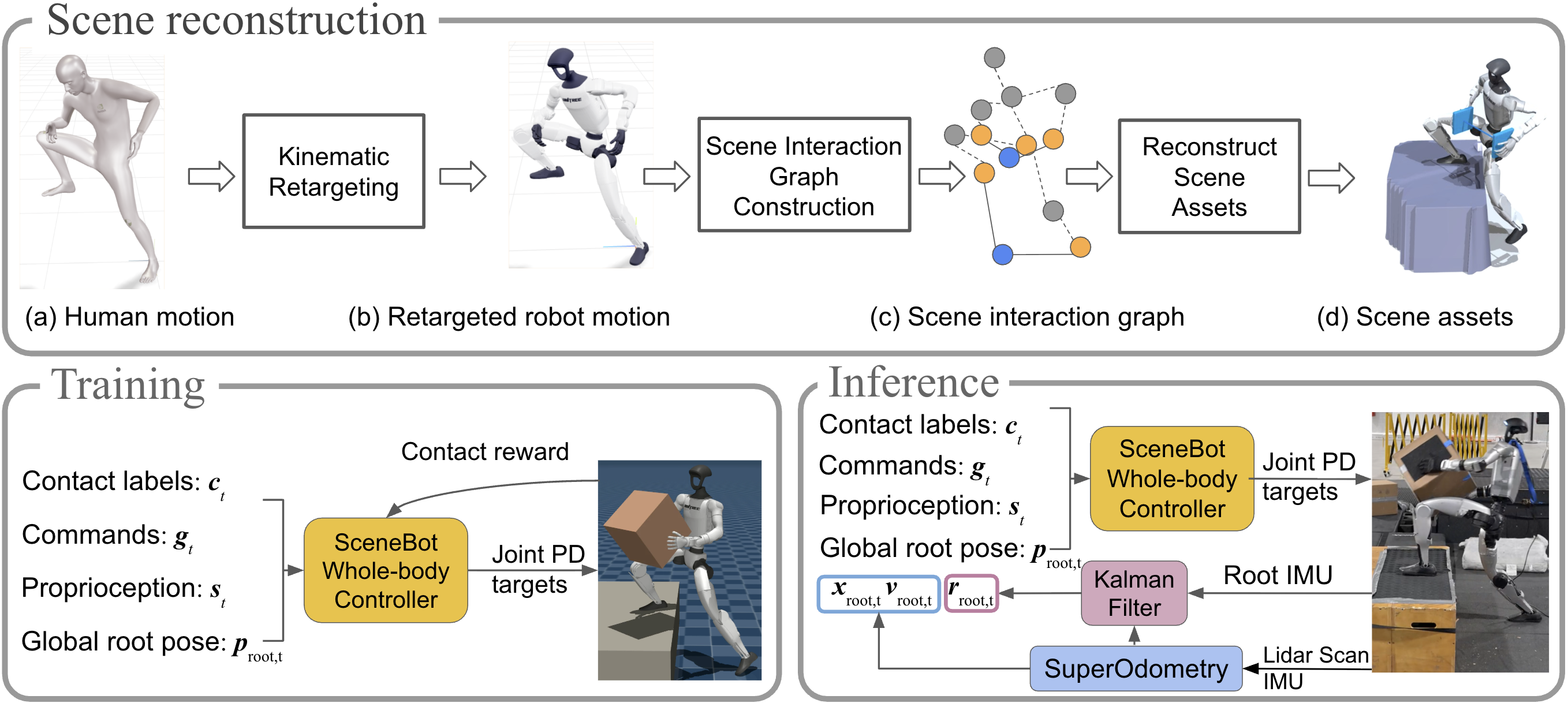}
  \caption{\textbf{Scene reconstruction:} SceneBot uses retargeted human motion to reconstruct scene assets. It first builds the robot-scene interaction graph, then reconstruct plausible terrains and objects. \textbf{Training:} SceneBot trains a motion and contact tracking policy via reinforcement learning using contact-based rewards. \textbf{Deployment:} SceneBot relies on SuperOdometry~\cite{superodom} and an onboard IMU to estimate root position $\bm{x}_\text{root,t}$, orientation $\bm{r}_\text{root,t}$ and linear veloity $\bm{v}_\text{root,t}$, enabling the robot to execute complex scene interactions, such as simultaneous terrain traversal and object carrying.}
  \label{fig:method}
  \vspace{-15pt}
\end{figure}

SceneBot is a framework designed to train a generalist tracking policy capable of robust terrain and object interaction. As illustrated in Fig.~\ref{fig:method}, SceneBot comprises several key components. Training a unified scene-interaction controller necessitates paired data linking robot motion with scene assets. To generate this training data, we introduce a scene reconstruction pipeline that synthesizes plausible terrains and objects directly from human kinematic motion. Using these reconstructed environments, we train a whole-body controller via reinforcement learning (RL) to track both the target kinematic motion and the desired contact labels. During deployment, this whole-body controller can successfully track reference motions with contact labels to facilitate scene interaction, as well as motions without contact labels for standard flat-terrain locomotion.

\subsection{Scene Generation from Human Motion}
\begin{wrapfigure}{r}{0.5\textwidth}
  \centering
  \includegraphics[width=\linewidth]{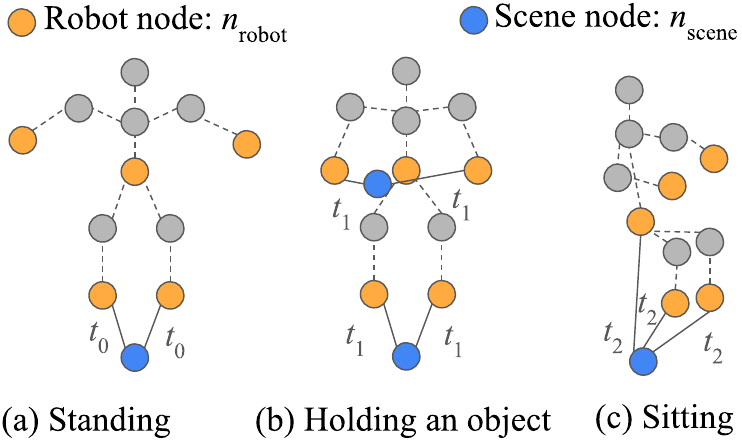}
  \caption{Scene interaction graph for different interactions.}
  \vspace{-15pt}
  \label{fig:interaction_graph}
\end{wrapfigure}
To facilitate training, paired robot reference motions and plausible scene assets (e.g., terrains and objects) are required. However, most existing datasets provide only human or robot kinematic motion absent of any scene context. Inspired by \cite{tip}, we reconstruct scene assets directly from motion. Given kinematically captured human motion, the scene reconstruction module first retargets it to the robot using kinematic motion retargeting \cite{gmr}. Scene reconstruction then proceeds in two stages: first, we construct a robot-scene interaction graph based on contact criteria and temporal consistency; second, we generate the terrain and object assets conditioned on this interaction graph.

\textbf{Scene Interaction Graph:} The scene interaction graph is a heterogeneous graph that captures the temporal interactions between the robot and the scene. As shown in Fig.~\ref{fig:interaction_graph}, the robot's key links are defined as $\mathcal{K} \in \{\text{left-wrist}, \text{right-wrist}, \text{left-foot}, \text{right-foot}, \text{pelvis}\}$. Each key link corresponds to a robot node $\mathcal{N}_\text{robot} = \{n^i_\text{robot} \mid i \in \mathcal{K}\}$ within the interaction graph. Two additional scene type nodes, $\mathcal{N}_\text{scene}=\{n^j_{\text{scene}}|j\in\{\text{terrain},\text{object}\}\}$, represent immovable terrain and movable objects, respectively. A temporal edge $\bm{e} : \{t_s, \{n^i_\text{robot}\}, \{n^j_\text{scene}\}, \bm{p}\}$ records an interaction between a robot link and the scene at a specific spatial location $\bm{p}$ and timestep $t_s$. The interaction graph is initially constructed from the robot's motion by adding candidate edges whenever the robot links exhibit low relative velocity and acceleration compared to the scene nodes, indicating potential contact. Infeasible edges are subsequently pruned if they result in collisions outside the designated interaction interval or if they violate force-closure constraints during object interactions.

\textbf{Scene Reconstruction:} After obtaining the scene interaction graph, terrains and objects are reconstructed based on its topological structure. Terrain is represented as a 2.5D elevation map. For each interaction edge associated with the terrain, a square plateau with height $h=p_z$ and center $(p_x,p_y)$ is added to the elevation map. Following the initial map construction, isolated plateaus of similar heights are merged, and regions that would cause robot-terrain collisions are carved out. Objects are represented as a set of plates parallel to the corresponding robot contact surfaces. The centers and trajectories of these objects are determined using the mean positions of the interaction edges involved in the grasping action. More detailed algorithm is described Apppendix.\ref{sec:scene_reconstruct}

\subsection{Training a Contact-Aware Whole-Body Controller}
Our whole-body tracking policy takes kinematic commands from the reference motion $\bm{g}_t \in\mathbb{R}^{93}$, desired contact labels $\bm{c}_t \in\{0,1\}^n$, and proprioceptive states $\bm{s}_t \in\mathbb{R}^n$ as input:
\[
\bm{a}_t=\pi(\bm{c}_t,\bm{g}_t,\bm{s}_t,\bm{p}_\text{root,t})
\]
The contact label is a binary vector representing all possible contact combinations between a robot link index $i \in \mathcal{K}$ and a contact scene type $j \in \{\text{terrain}, \text{object}\}$, formulated as $\bm{c}_t = [c^{i,j}_t]$. An active contact $c^{i,j}_t = 1$ indicates that robot node $i$ should establish contact with scene type $j$. The policy command $\bm{g}_t$ includes the desired joint angles and velocities for the lower body, the 6D poses of the head and wrists in the robot's root frame, and the global root position and orientation errors between the current and target root states, following the convention established by \cite{beyondmimic,chip}. The proprioceptive states $\bm{s}_t$ consist of the robot's current joint angles and velocities, the projected gravity in the root frame, and the root angular velocity. For root state $\bm{p}_\text{root,t}$, it include root position, root orientation and root linear velocity.

We train our whole-body control policy using the Proximal Policy Optimization (PPO) reinforcement learning algorithm to simultaneously track the kinematic reference motion and the desired contact behavior. We utilize motion tracking rewards similar to those in \cite{beyondmimic, sonic}, augmented by two contact-specific reward terms. First, we introduce a contact correctness reward $r_\text{cr}$ to encourage accurate contacts:
\begin{align}
    r_\text{cr} &= \sum_i(c^{i,j}_{\text{des},t} \equiv c^{i,j}_{\text{actual},t}), \quad \mathrm{if\;} c^{i,j}_{\text{des},t} = 1, 
\end{align}
where $\bm{c}_{\text{des},t}$ and $\bm{c}_{\text{actual},t}$ denote the desired contact labels and the actual contact labels detected in the simulator, respectively. To leverage a large corpus of flat-terrain motions lacking contact labels, the terrain contact reward is considered only when the desired contact label is set to 1. This asymmetric design ensures that an unintentional or unspecified contact event would not be penalized.  During deployment, our policy robustly tracks flat-terrain free-form motions by simply zeroing out the contact labels.

In addition, we introduce a contact duration reward $r_\text{dr}$ to encourage stable contacts:
\begin{align}
    r_\text{dr} &= \sum_i \min(t^{i,j}_c, 0.5), \quad i \in \mathcal{K}, \; j \in \{\text{terrain,\;object}\},
\end{align}
where $t^{i,j}_c$ represents the cumulative contact time for a specific interaction. Contact duration reward is clipped at 0.5 second to prevent robot from refusing to release the object.

For object interactions such as grasping, achieving the desired contact label promptly is crucial for securing the object. To enforce this, we introduce a contact-mismatch termination condition that enforces the policy to establish the desired contact within a specified time frame. 

Since our object motion is also reconstructed from only robot motion, not all objects start in a resting position on the floor prior to grasping; therefore, we apply a heuristic stabilizing force (a ``magic force'') to the object, ensuring it follows the reconstructed trajectory until a stable force-closure grasp is achieved. This intervention mitigates frequent grasp-failure terminations during the early stages of training, thereby providing a denser contact reward signal.

\subsection{Infrastructure-Free State Estimation} 

We employ SuperOdometry \cite{superodom}, a state-of-the-art LiDAR-inertial odometry framework, to estimate the robot's global root position, orientation, and linear velocity. SuperOdometry provides robust and accurate state estimation across diverse environments and dynamic motions. However, during humanoid deployment, we found that relying solely on the orientation estimated from the Livox IMU is insufficient for accurately recovering the humanoid root orientation. This behavior mainly arises from the unique sensor configuration of the humanoid platform. The Livox IMU is mounted upside down on the robot's head, and its motion and orientation do not always accurately represent the true dynamics of the humanoid root body. During aggressive motions, head movement can deviate substantially from pelvis dynamics, making the estimated orientation less representative of the root state required for stable control. To obtain a more representative estimate of the robot root state, we directly compute the pitch and roll angles from a pelvis-mounted root IMU. For yaw estimation, we apply a Kalman filter to fuse the root IMU yaw angular velocity with the yaw orientation estimated by SuperOdometry. The complete state estimation pipeline operates at 200 Hz. During initialization, the root height is estimated using the lowest 5\% of points from the gravity-aligned LiDAR scan, while the world-frame origin is initialized at the robot's starting $x$ and $y$ coordinates with zero height.

\section{Experiments}
\label{sec:experiment}
In this section, we will first show the tracking performance of our model, then analyze the effects of key design choices in our method. We will also discuss the comparison between scene-aware retargeting and scene reconstruction, as well as the accuracy of root state estimation.


\begin{figure}[h]
  \centering
  \includegraphics[width=\linewidth]{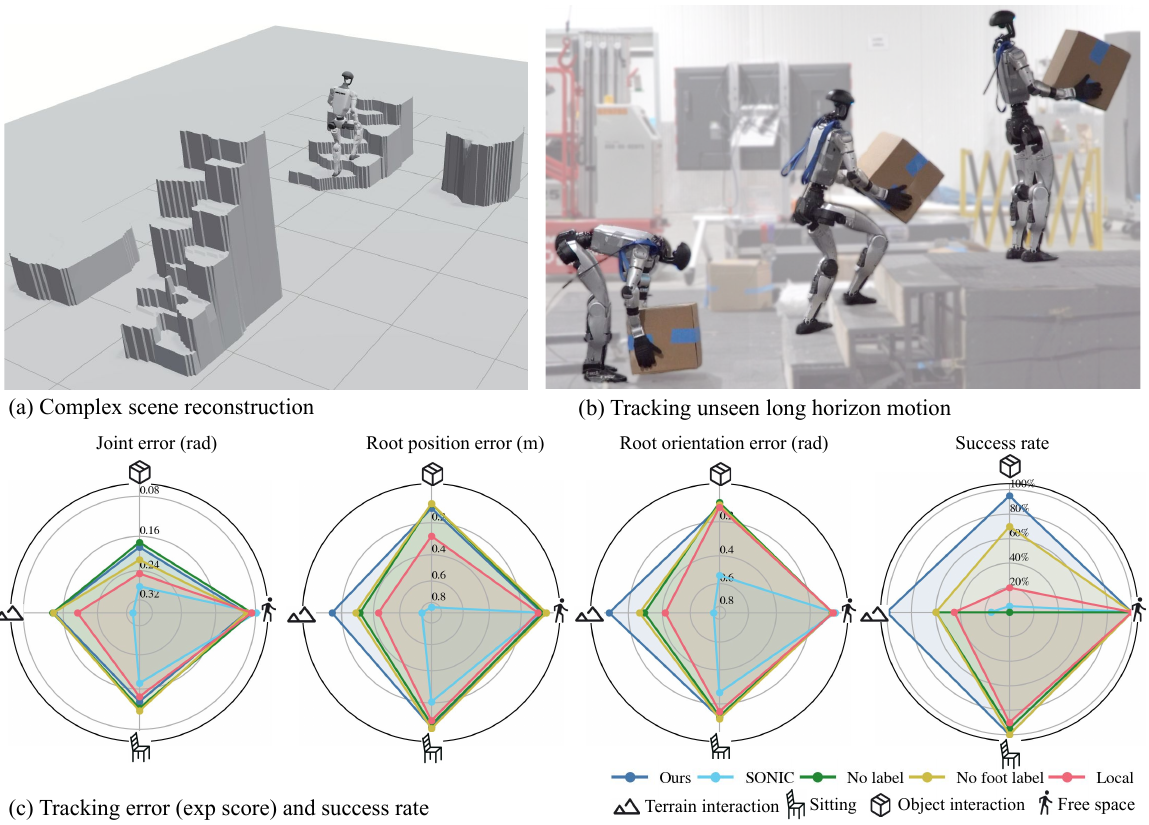}
  \caption{(a): Our scene reconstruction method can generate complex scenes that match the retargeted motion. (b):The general motion tracking controller can track long-horizon scene interaction motions that require simultaneous object manipulation and terrain traversal. (c): We use negative exponential scaling for error plot, larger contour area indicates better performance.}
  \label{fig:main_result}
  \vspace{-15pt}
\end{figure}

\subsection{Experiment Setup and Dataset}
\begin{wrapfigure}{r}{0.4\textwidth}
  \centering
  \includegraphics[width=\linewidth]{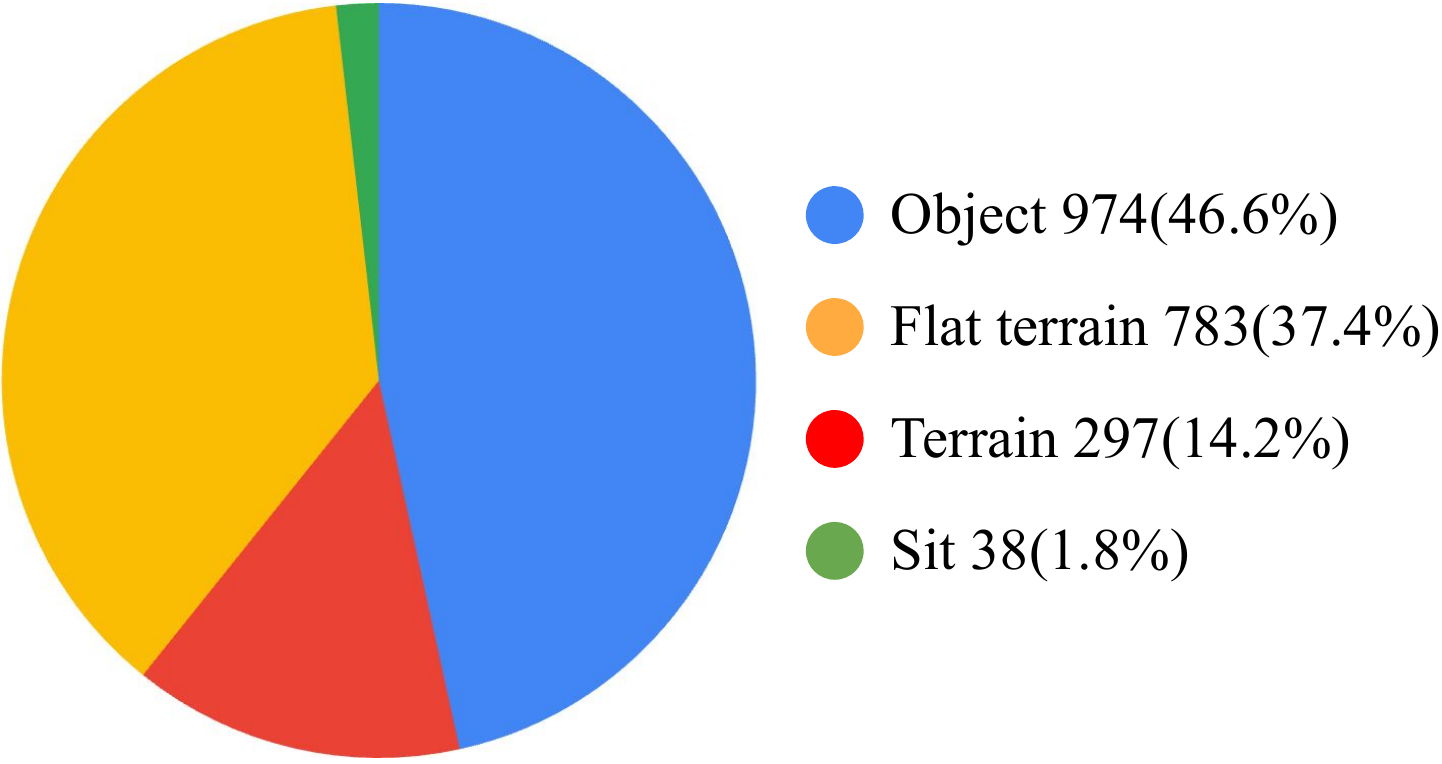}
  \caption{Composition of training data.}
  \vspace{-15pt}
  \label{fig:dataset}
\end{wrapfigure}

We use the Unitree G1 as our testing platform. To execute the policy and perform state estimation, the robot is connected via an Ethernet cable to a laptop equipped with an Intel i5-13200H CPU and an NVIDIA RTX 3050 GPU. For state estimation, we utilize the G1's built-in LiDAR and the IMU sensor located within its pelvis. To compile our training data, we combine AMASS \cite{amass}, OMOMO \cite{omomo}, Bones \cite{sonic}, and Lafan \cite{lafan} datasets, providing a diverse mixture of scene interaction and flat-terrain movement data. The overall dataset distribution is illustrated in Fig. \ref{fig:dataset}.

\subsection{Tracking Performance for Different Motions}
\begin{wrapfigure}{r}{0.3\textwidth}
  \centering
  \includegraphics[width=\linewidth]{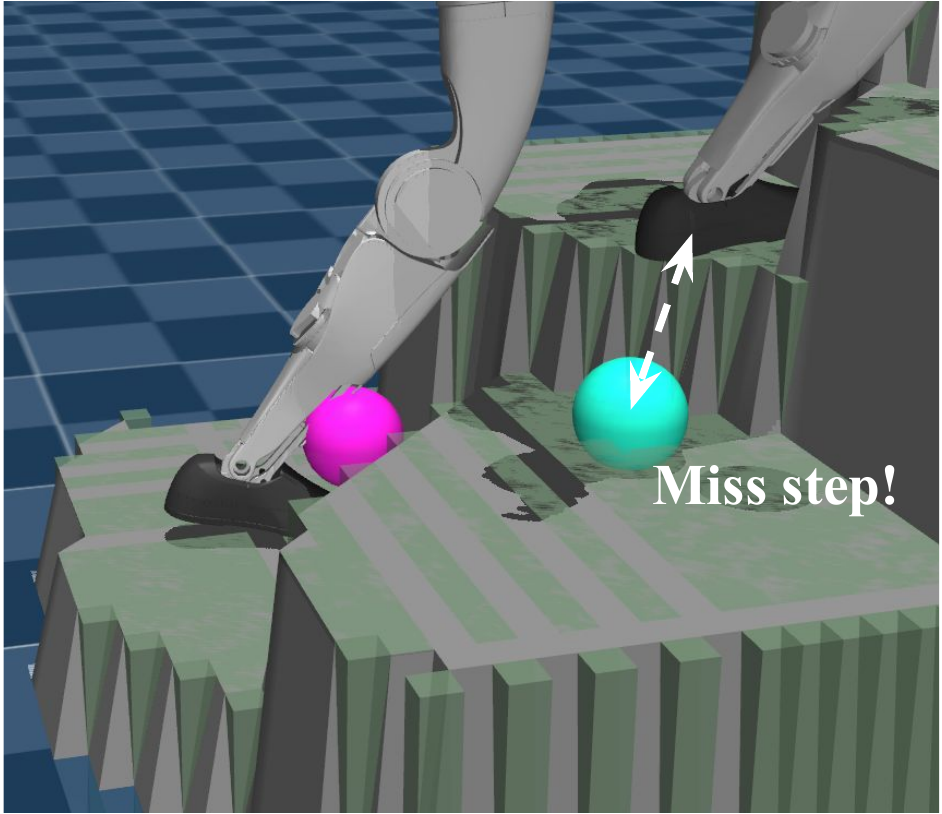}
  \caption{Drift in local tracking causes motion-terrain misalignment.}
  \vspace{-15pt}
  \label{fig:failure_misalign}
\end{wrapfigure}
Qualitatively, our method successfully executes behaviors such as stepping onto stairs of varying heights, picking up and carrying boxes, sitting down, and performing agile kicking and running, as demonstrated in the supplementary video. Additionally, our approach manages long-horizon, simultaneous object and terrain interactions, such as carrying a box while ascending stairs (Fig. \ref{fig:main_result}).

Quantitatively, we evaluate tracking performance using the average global root tracking error, average joint tracking error, and success rates across four task categories in a MuJoCo sim-to-sim environment: free-space, terrain interaction, object interaction, sitting. We compare our method against the state-of-the-art general motion tracking policy, SONIC \cite{sonic} as well as baselines trained without contact labels (No label), without foot contact labels but with hand labels (No foot label), as well as without global root information (Local). For each category, we use 20 unseen motion sequences paired with scene assets generated by our scene reconstruction pipeline. As shown in Fig. \ref{fig:main_result}, our model achieves performance on par with SONIC for free-space motion tracking, while significantly outperforming it on all scene interaction tasks.n For different type of tasks, success definition can be found in Appendix.\ref{sec:success}

\subsection{Necessity of Global Tracking}
\begin{wrapfigure}{r}{0.3\textwidth}
  \centering
  \includegraphics[width=\linewidth]{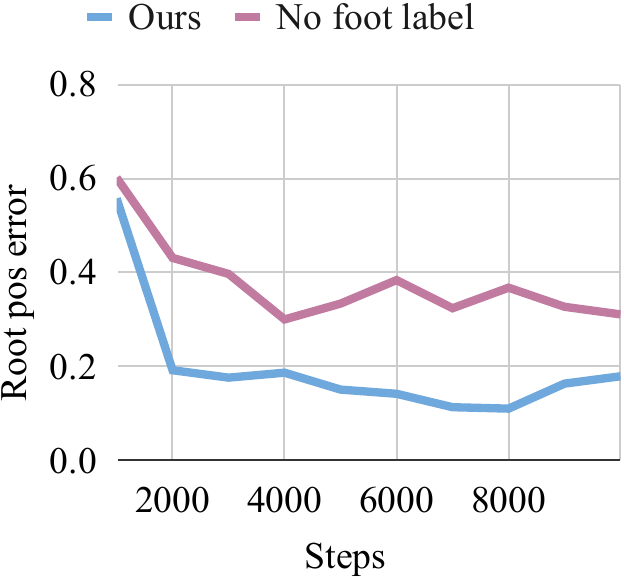}
  \vspace{-10pt}
  \caption{Root position error on the terrain task across different training steps.}
  \vspace{-15pt}
  \label{fig:training_label}
\end{wrapfigure}
As illustrated in Fig. \ref{fig:main_result}, training the whole-body tracking policy without global root position and linear velocity results in significantly worse tracking performance compared to our method using global state estimation (Terrain task: $100\% \rightarrow 45\%$ and object-related tasks $95\% \rightarrow 20\%$). Incorporating global root positioning prevents drift during both training and inference. It enables the policy to learn faster by reducing the number of misalignment-related terminations during training and avoids misalignment-induced failures during testing (Fig. \ref{fig:failure_misalign}).

\subsection{Effectiveness of Contact Labels}

For object grasping, contact labels play an essential role in task success. If the policy is trained without hand contact labels, or if they are deactivated, the object grasp success rate drops significantly to near zero (Fig. \ref{fig:main_result}c, No label). The robot subsequently fails to pick up or stably hold the object. For terrain traversal, contact labels are particularly helpful during frequent contact switches, such as stepping on or off stairs. A model trained without foot contact labels suffers from missteps after initial movements, ultimately resulting in the robot falling over. 

Interestingly, disabling contact labels only during testing still yields better performance (an 85\% success rate, just 15\% lower than when using active contact labels) compared to a model explicitly trained without them on terrain traversal environments (No foot label). We hypothesize that incorporating contact labels simplifies the training process; the learned balancing mechanics are retained even when explicit contact labels are omitted during inference, as supported by Fig. \ref{fig:training_label}.

\subsection{Retargeting vs. Reconstruction}
\begin{wrapfigure}{r}{0.5\textwidth}
  \centering
  \includegraphics[width=\linewidth]{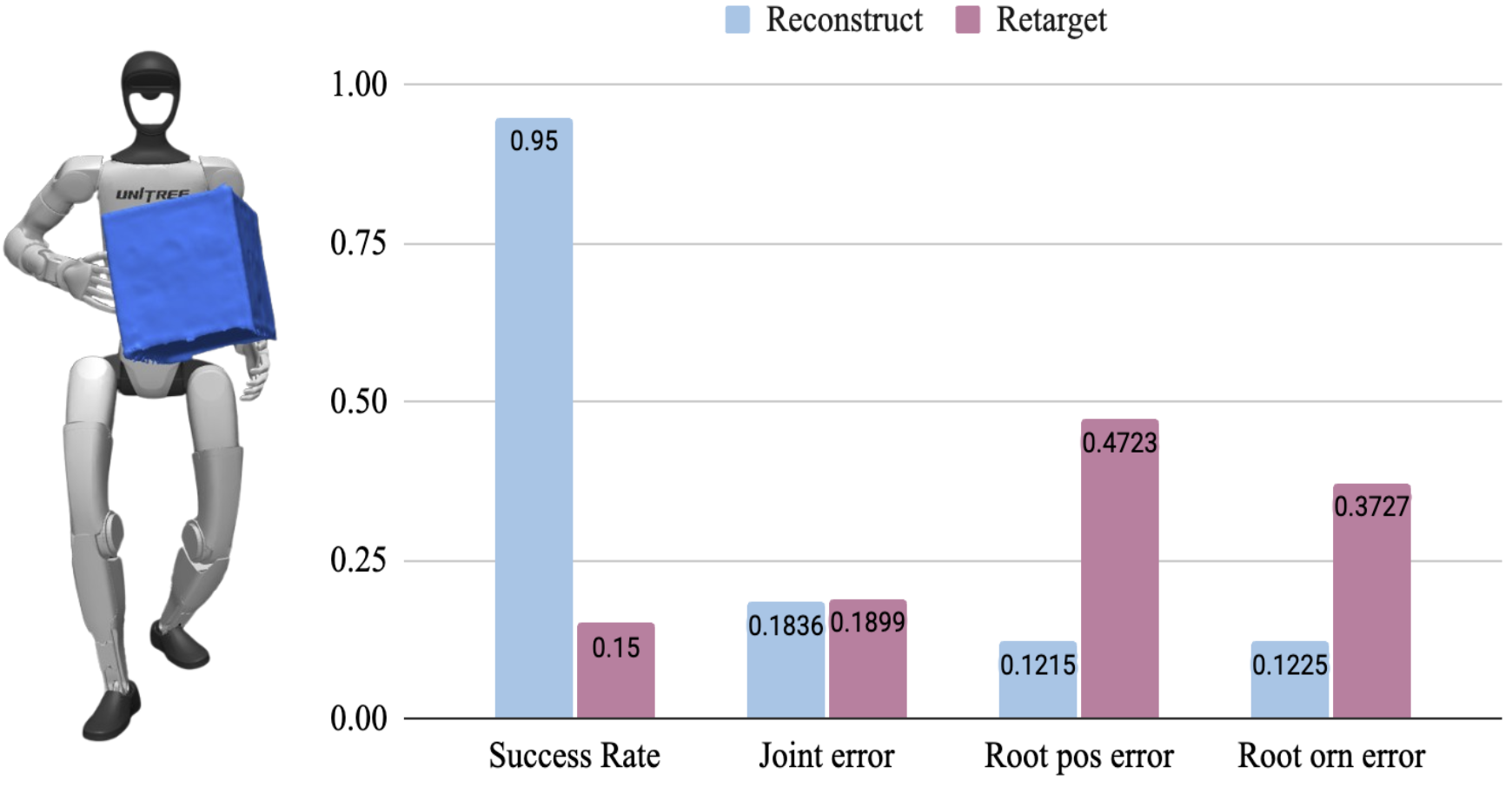}
  \caption{Left: Omni-retarget creates an object-hand mismatch. Right: Tracking performance comparison between scene asset reconstruction and scene-aware retargeting.}
  \vspace{-10pt}
  \label{fig:retarget_reconstruct}
\end{wrapfigure}
Qualitatively, our pipeline can reconstructs complicate terrain from Lafan obstacle sequences (Fig. \ref{fig:main_result}) and object assets from Bones and OMOMO interactions. Compared to scene-ware retargeting, our method guarantees that the generated scene is feasible for supporting robot motion instead of prioritizing obeying constraints on the existing scene. This advantage could benefit general motion tracking training as there is less load for RL to fix the robot motion and scene mismatch. To verify our hypothesis, we compared our direct scene reconstruction against a tuned Omni-retarget baseline on the OMOMO dataset. As shown in Fig. \ref{fig:retarget_reconstruct}, training with reconstructed scene outperforms using Omni-retarget. One example we notice is that retargeting struggles to perfectly align objects while avoiding penetration; the resulting policy suffers from higher grasp failure rates.

\subsection{Performance of State Estimation}
\begin{wrapfigure}{r}{0.55\textwidth}
  \centering
  \includegraphics[width=\linewidth]{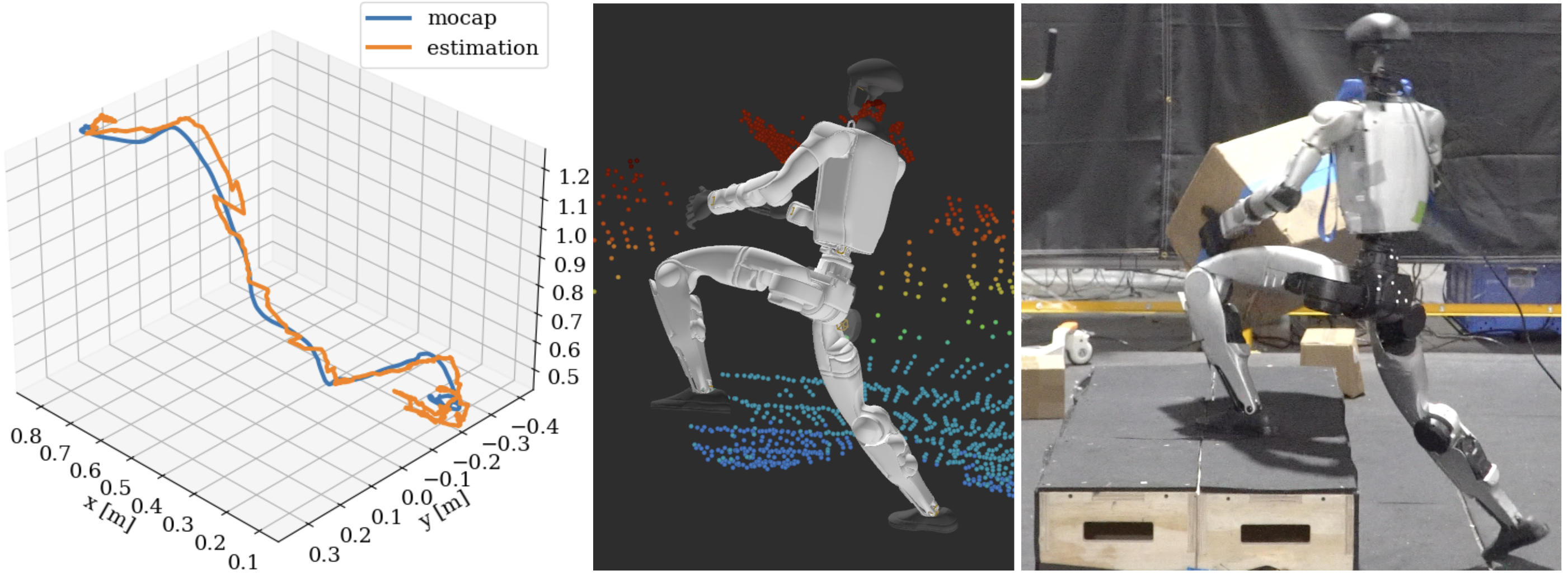}
  \caption{Left: State estimation compared against mocap ground truth. Middle: Point cloud visualization. Right: Real world image.}
  \vspace{-10pt}
  \label{fig:odom_result}
\end{wrapfigure}
We evaluate our state estimation system by comparing the estimated root pose against the ground truth from an OptiTrack mocap system. We measure the results using challenging motions, such as picking up a box and stepping onto a stage, as shown in Fig. \ref{fig:odom_result}. Overall, the system achieves an average root position error of 0.032 m, a root orientation error of 0.018 rad, and a root linear velocity error of 0.092 m/s, alongside an 80\% task success rate across 10 experiments. The comparision with the mocap ground truth is shown in Fig. \ref{fig:odom_result}, more result can be found in Appendix.\ref{sec:state_estimation}.

\section{Limitations}
\label{sec:limitation}
Successful scene reconstruction requires high-fidelity robot motion and is highly sensitive to poorly retargeted human motions, such as foot sliding. Additionally, motion trajectories extracted from raw videos or generative models often lack physical consistency, which risks destabilizing our scene interaction graph construction.

\section{Conclusion}
\label{sec:conclusion}
We introduced Scenebot, a contact-prompted whole-body tracking policy, alongside a data engine that synthesizes interactive scenes from robot motion for large-scale training. Evaluations demonstrate that Scenebot accurately tracks diverse behaviors---ranging from flat-terrain locomotion to complex, multi-task environmental interactions, such as ascending stairs while carrying an object. Future work aims to develop an autonomous high-level controller for perceptive, goal-driven scene interactions.


\clearpage
\acknowledgments{}


\bibliography{reference}  
\newpage
\appendix
\section{Scene Reconstruction Algorithm}
\label{sec:scene_reconstruct}
\begin{algorithm}
\caption{Scene Reconstruction from Human Motion}
\label{alg:scene_reconstruction}
\begin{algorithmic}[1]
\Require Human kinematic motion $M_{\text{human}}$
\Ensure Reconstructed Terrain $T$, Reconstructed Objects $O$, Robot Motion $M_{\text{robot}}$, Interaction Graph $G$

\Statex \textbf{Stage 1: Motion Retargeting}
\State $M_{\text{robot}} \gets \text{KinematicRetargeting}(M_{\text{human}})$

\Statex \textbf{Stage 2: Construct Scene Interaction Graph}
\State Let $\mathcal{K} = \{\text{left-wrist}, \text{right-wrist}, \text{left-foot}, \text{right-foot}, \text{pelvis}\}$
\State Initialize robot nodes $\mathcal{N}_{\text{robot}} = \{n^i_{\text{robot}} \mid i \in \mathcal{K}\}$
\State Initialize scene nodes $\mathcal{N}_{\text{scene}} = \{\text{terrain}, \text{object}\}$
\State Initialize edge set $\mathcal{E} \gets \emptyset$

\For{each timestep $t_s$ in $M_{\text{robot}}$}
    \For{each key link $i \in \mathcal{K}$}
        \If{velocity and acceleration of $i$ relative to scene are low}
            \State Determine interaction location $\mathbf{p}$ and target scene node $n^j_{\text{scene}}$
            \State Add candidate edge $\mathbf{e} = \{t_s, \{n^i_{\text{robot}}\}, \{n^j_{\text{scene}}\}, \mathbf{p}\}$ to $\mathcal{E}$
        \EndIf
    \EndFor
\EndFor

\Statex \Comment{Prune infeasible edges}
\For{each edge $\mathbf{e} \in \mathcal{E}$}
    \If{causes collision outside interaction interval \textbf{or} violates force-closure constraints}
        \State $\mathcal{E} \gets \mathcal{E} \setminus \{\mathbf{e}\}$
    \EndIf
\EndFor
\State $G \gets (\mathcal{N}_{\text{robot}} \cup \mathcal{N}_{\text{scene}}, \mathcal{E})$

\Statex \textbf{Stage 3: Scene Reconstruction}
\State Initialize 2.5D elevation map $T$ for terrain
\State Initialize object set $O$

\Statex \Comment{Reconstruct Terrain}
\For{each $\mathbf{e} \in \mathcal{E}$ connected to $n^{\text{terrain}}_{\text{scene}}$}
    \State Extract spatial location $\mathbf{p} = (p_x, p_y, p_z)$ from $\mathbf{e}$
    \State Add square plateau at center $(p_x, p_y)$ with height $h = p_z$ to $T$
\EndFor
\State Merge isolated plateaus of similar heights in $T$
\State Carve out regions in $T$ that cause robot-terrain collisions

\Statex \Comment{Reconstruct Objects}
\For{each grasping action grouped from edges $\mathbf{e} \in \mathcal{E}$ connected to $n^{\text{object}}_{\text{scene}}$}
    \State Generate plates parallel to corresponding robot contact surfaces
    \State Compute object center and trajectory using mean positions of interaction edges
    \State Assemble plates and trajectory into an object and add to $O$
\EndFor

\State \Return $T, O, M_{\text{robot}}, G$
\end{algorithmic}
\end{algorithm}

\section{Training rewards}
\label{sec:rewards}
Detailed tracking rewards are defined in Tab.\ref{tab:rl_rewards}
\begin{table}[h]
\centering
\setlength{\tabcolsep}{2pt} 
\renewcommand{\arraystretch}{0.85} 
\begin{tabular}{l l c}
\toprule
\textbf{Reward term} & \textbf{Equation} & \textbf{Weight} \\
\midrule
\multicolumn{3}{l}{\textit{Tracking rewards}} \\
Root orientation & $r^{\text{root}}_{\text{ori}}(t)=\exp\!\big(- \|\bm{r}^p_{t,r}-\bm{r}^g_{t,r}\|_2^2 / 0.4^2\big)$ & 0.5 \\
Root position & $r^{\text{root}}_{\text{ori}}(t)=\exp\!\big(- \|\bm{x}^p_{t,r}-\bm{x}^g_{t,r}\|_2^2 / 0.3^2\big)$ & 0.5 \\
Body link pos (rel.) & $r^{\text{body}}_{\text{pos}}(t)=\exp\!\Big(-\tfrac{1}{|\mathcal{B}|}\!\sum_{b\in\mathcal{B}}\|\bm{p}^{p,\text{rel}}_{t,b}-\bm{p}^{g,\text{rel}}_{t,b}\|_2^2 / 0.3^2\Big)$ & 1.0 \\
Body link ori (rel.) & $r^{\text{body}}_{\text{ori}}(t)=\exp\!\Big(-\tfrac{1}{|\mathcal{B}|}\!\sum_{b\in\mathcal{B}} \|\bm{o}^{p,\text{rel}}_{t,b}-\bm{o}^{g,\text{rel}}_{t,b}\|_2^2 / 0.4^2\Big)$ & 1.0 \\
Body link lin. vel & $r^{\text{body}}_{\text{lin}}(t)=\exp\!\Big(-\tfrac{1}{|\mathcal{B}|}\!\sum_{b\in\mathcal{B}}\|\bm{v}^p_{t,b}-\bm{v}^g_{t,b}\|_2^2 / 1.0^2\Big)$ & 1.0 \\
Body link ang. vel & $r^{\text{body}}_{\text{ang}}(t)=\exp\!\Big(-\tfrac{1}{|\mathcal{B}|}\!\sum_{b\in\mathcal{B}}\|\bm{\omega}^p_{t,b}-\bm{\omega}^g_{t,b}\|_2^2 / 3.14^2\Big)$ & 1.0 \\
\\
\multicolumn{3}{l}{\textit{Contact rewards}} \\
Contact label reward & $r_\text{cr}(t) = \sum_i(c^{i,j}_{\text{des},t} \equiv c^{i,j}_{\text{actual},t}), \quad \mathrm{if\;} c^{i,j}_{\text{des},t} = 1$ & 0.5 \\
Contact duration reward & $r_\text{dr}(t) = \sum_i \min(t^{i,j}_c, 0.5), \quad i \in \mathcal{K}, \; j \in \{\text{terrain,object}\}$ & 0.5 \\
\\
\multicolumn{3}{l}{\textit{Penalty terms $\mathcal{P}({\bm{s}^{\text{p}}_t}, \bm{a}_t)$}} \\
Action rate & $r_{\text{act}}(t)=\|\bm{a}_t-\bm{a}_{t-1}\|_2^2$ & -0.1 \\
Joint limit & $r_{\text{jlim}}(t)=\sum_{j} \bm{1}[\bm{q}_{t,j} \notin [\bm{q}_{t,j}^{\text{min}}, \bm{q}_{t,j}^{\text{max}}]]$ & -10.0 \\
Undesired contacts & $r_{\text{contact}}(t)=\sum_{c \notin \{\text{ankles, wrists}\}} \bm{1}[\|\bm{F}_c\|>1.0~\text{N}]$ & -0.1 \\
\bottomrule
\end{tabular}
\caption{Reward design details. $r$: root, $b\in\mathcal{B}$: body links.}
\label{tab:rl_rewards}
\end{table}

\section{Quantitative results}
\label{sec:quan_results}
For each quantitative value in Fig.\ref{fig:main_result}, such as joint tracking error, root position and orientation tracking error and success rate, detailed value can be found in Tab.\ref{tab:Quantitative results}
\begin{table}[htbp]
  \centering
  \caption{Evaluation metrics across different methods and scenarios.}
  \label{tab:evaluation_results}
  \begin{tabular}{l c c c c}
    \toprule
    \textbf{Method} & \textbf{free} & \textbf{object} & \textbf{terrain} & \textbf{sit} \\
    \midrule
    
    \multicolumn{5}{c}{\textbf{Joint error (rad)}} \\
    \midrule
    Ours                & 0.0977 & 0.1836 & 0.1406 & 0.1335 \\
    SONIC               & 0.0784 & 0.2814 & 0.3390 & 0.1728 \\
    No label            & 0.0874 & 0.1729 & 0.1378 & 0.1204 \\
    No foot label       & 0.0976 & 0.2133 & 0.1392 & 0.1148 \\
    Local               & 0.0876 & 0.2471 & 0.1918 & 0.1441 \\
    Ours(No foot label) & -      & -      & 0.1363 & 0.1413 \\
    Syn Only            & -      & 0.1899 & -      & -      \\
    \midrule
    
    \multicolumn{5}{c}{\textbf{Root position error (m)}} \\
    \midrule
    Ours                & 0.0893 & 0.1215 & 0.1471   & 0.0702 \\
    SONIC               & 0.0973 & 0.8491 & 0.808002 & 0.2051 \\
    No label            & 0.0812 & 0.0979 & 0.3103   & 0.0840 \\
    No foot label       & 0.0661 & 0.0970 & 0.2834   & 0.0636 \\
    Local               & 0.1091 & 0.2782 & 0.4344   & 0.1043 \\
    Ours(No foot label) & -        & -    & 0.1562   & 0.0726  \\
    Syn Only            & -      & 0.4723 & -        & -      \\
    \midrule
    
    \multicolumn{5}{c}{\textbf{root orientation error (rad)}} \\
    \midrule
    Ours                & 0.0854 & 0.1225 & 0.1017 & 0.1299 \\
    SONIC               & 0.0748 & 0.5354 & 0.8046 & 0.2620 \\
    No label            & 0.0888 & 0.1031 & 0.2890 & 0.1363 \\
    No foot label       & 0.0883 & 0.1153 & 0.2609 & 0.1234 \\
    Local               & 0.0849 & 0.1262 & 0.4178 & 0.1603 \\
    Ours(No foot label) & -      & -      & 0.1190 & 0.1623 \\
    Syn Only            & -      & 0.3727 & -      & -      \\
    \midrule
    
    \multicolumn{5}{c}{\textbf{Success rate}} \\
    \midrule
    Ours                & 100.00\% & 95.00\% & 100.00\% & 100.00\% \\
    SONIC               & 100.00\% & 5.00\%  & 15.00\%  & 0.00\%   \\
    No label            & 100.00\% & 0.00\%  & 60.00\%  & 95.00\%  \\
    No foot label       & 100.00\% & 70.00\% & 60.00\%  & 100.00\% \\
    Local               & 100.00\% & 20.00\% & 45.00\%  & 90.00\%  \\
    Ours(No foot label) & -        & -       & 85.00\%  & 100.00\% \\
    Syn Only            & -        & 15.00\% & -        & -        \\
    \bottomrule
  \end{tabular}
  \label{tab:Quantitative results}
\end{table}
\section{Definition of Success}
\label{sec:success}
\begin{itemize}
\item \textbf{Terrain}: The robot doesn't fall down; root height error less than 0.2m.
\item \textbf{Object}: The robot doesn't fall down; the object is grasped until release.
\item \textbf{Sit}: The robot doesn't fall down; the robot's pelvis makes contact with the chair.
\item \textbf{Free space}: The robot doesn't fall down.
\end{itemize}
\section{State estimation results}
\label{sec:state_estimation}

\begin{figure}[h]
  \centering
  \includegraphics[width=\linewidth]{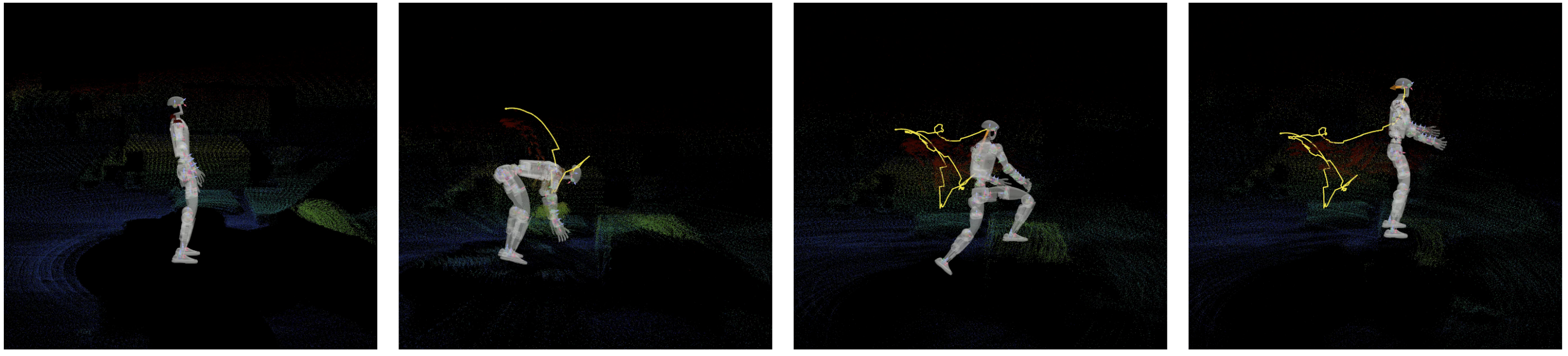}
  \caption{Detailed breakdown of the state estimation results for the task of grasping a box and stepping onto a stage. The yellow line visualizes the estimated trajectory, while the point cloud represents the map built using SuperOdometry.}
  \label{fig:breakdown_sa}
  \vspace{-15pt}
\end{figure}

Figure \ref{fig:breakdown_sa} shows keyframes of the state estimation results. The maximum error occurs when the robot crouches down. This sudden crouching motion generates a scan that differs significantly from the reference point map, temporarily increasing the state estimation error. The state estimation accuracy subsequently recovers as the robot straightens up.

\end{document}